\documentclass[10pt,twocolumn,letterpaper]{article}

\usepackage{cvpr}
\usepackage{times}
\usepackage{epsfig}
\usepackage{graphicx}
\usepackage{amsmath}
\usepackage{amssymb}
\usepackage{graphicx}
\usepackage{subfigure}
\usepackage{imakeidx}

\usepackage[breaklinks=true,bookmarks=false]{hyperref}

\cvprfinalcopy % *** Uncomment this line for the final submission

\def\cvprPaperID{****} % *** Enter the CVPR Paper ID here

\begin{document}

%%%%%%%%% TITLE
% \title{An integrated approach on cell segmentation of phase contrast images}

% {\tt\small firstauthor@i2.org}
% \and
% \and
% Third Author\\
% Institution3\\
% First line of institution3 address\\
% {\tt\small secondauthor@i3.org}
% }

% \maketitle
% %\thispagestyle{empty}

\title{An Efficient Approach for  Cell Segmentation in Phase Contrast Microscopy Images}

\author{Lin Zhang\\
%Department of Computer Science\\
{\tt\small lzhang22@albany.edu}
\and
% Mei Chen\\
% %Department of Computer Engineering\\
% {\tt\small meichen@albany.edu}
 }

\maketitle

%%%%%%%%% ABSTRACT
\begin{abstract}
In this paper, we propose a new model to segment cells in phase contrast microscopy images. Cell images collected from the similar scenario share a similar background. Inspired by this, we separate cells from the background in images by formulating the problem as a low-rank and structured sparse matrix decomposition problem. Then, we propose the inverse diffraction pattern filtering method to further segment individual cells in the  images. This is a deconvolution process that has a much lower computational complexity when compared to the other restoration methods. Experiments demonstrate the effectiveness of the proposed model when it is compared with recent works.
\end{abstract}

%%%%%%%%% BODY TEXT
\section{Introduction}
Phase contrast microscopy is one of the most fundamental invention that we can observe cells without any damages. 
Tremendous amounts of images are obtained from 
the phase contrast microscopy and the analysis of these images is critical. Cell segmentation in phase contrast microscopy images is still a great challenge because of the diversity of cells' appearance and artifacts. Many approaches have been proposed to deal with these issues. By taking advantage of the  gradient of cells' boundary, methods like Active Contour~\cite{Snakes} is introduced. However, the boundary cues are not always strong enough, therefore, this method often either can not converge or over-segment. 
To learning a statistical model from data, machine learning based methods~\cite{conf/miccai/FunkeHZ15,5540037} are proposed.
% Based on challenges mentioned earlier, these assumptions are not always held because of low contrast boundary between cells, and also low contrast intensity between cells and background.
However, these generic image analysis techniques have limits in the case of 
cell adhesion, cell event, artifatcs, etc.
% features between cells and background, cells and cells, and background and background are computed for identifying each individual cell. This method achieved good results on data with clear boundary. While, extracting these features is challenge when differences between cells, cells and background are not clear. 
It is evident that generic image analysis techniques, which are designed for natural images, have limits in phase contrast microscopy cell segmentation problem. 
Until recently, Yin \etal~\cite{understanding,Zhaozheng} analyzed the uniqueness of the phase  contrast  microscopy and proposed a method via modeling phase contrast imaging theory. Though effective, these methods have a huge computation cost. 
Another novel method~\cite{cellsensitive} proposed by modeling the differences between cells and non-cells when absorbing lights. This method needs sufficient prior information to tuning the parameters.

Background subtraction is an important pre-processing step in phase contrast microscopy cell segmentation, however limited number of works have done this. One common method is rolling ball filtering~\cite{Li2007}. But this method fails to work if the background is not uniform, which is quite often in phase contrast images. Another common method is to model the background as a second-order polynomial function~\cite{Zhaozheng,understanding}. However, the background in phase contrast microscopy images is more complicated than that, hence this model can not remove the background sufficiently. In recent years, low-rank techniques have been introduced to deal with background subtraction in natural images. Wright \etal~\cite{NIPS2009_3704} modelled the background as a low-rank matrix approximately and the foreground image as a sparse one. This method can learn the model of background from data directly.

\indent In this work, we utilize the fact that images obtained from the similar environment tend to have similar background. Therefore cells can be obtained by subtracting the background. We formulate this problem as a low-rank and structured sparse matrix decomposition problem. Due to the existence of noise, we propose inverse diffraction pattern filtering to get accurate individual cells.

%This model neglects the prior information of the structure from the one of the foreground. To address this problem, a model integrated with structural prior information of the foreground objects was recently introduced in~\cite{DBLP:journals/corr/XinTWG15}. The structural prior information extraction method was developed for natural images. The aforementioned imaging principle of phase contrast images is drastically different from that of nature images, which results in a significant difference between their structures. Thus, it's necessary to introduce a new way to extract structural prior information in phase contrast microscopy images.\\
%, where the information of phase retardation reflects the intrinsic structure of cells.

% , which can be efficiently solved in closed form.
%-------------------------------------------------------------------------

\section{Methodology}
%In order to segment cells in phase contrast microscopy images, we propose two 
% In this section, we illustrate proposed method for background subtraction firstly. To subtract background in phase contrast microscopy images, we introduce low-rank and structured sparse matrix decomposition to . 
% Then, we restore artifacts-free phase contrast microscopy images by proposed inverse diffraction patterns. 
In this section, we first illustrate the proposed method for background subtraction, which is based on low-rank and structured sparse matrix decomposition. Then, we demonstrate our proposed inverse diffraction patterns filtering.

\subsection{Cell Images Background Subtraction}
For a phase contrast microscopy image sequence, 
scenario that defined as  { \bf A} = [ $a_1$, $a_2$, $a_3$, {$\dots$}, $a_n$]. %where each frame is  $a_i$ \in ${\mathbb R^{w \times h}}$. 
For each microscopy image, we can model it as the combination of a background part and a foreground part as {$a_i = b_i+e_i$},
where $b_{i}$ and $e_{i}$ are the matrices for the background and foreground, respectively. It is easy to find that the background of images from a certain image sequence are linearly correlated. 
% From the imaging condition, the medium used to cultivate cells and illumination of the images undergoes very little changes for a certain image sequence. 
% Hence, these background of images from a certain image sequence are linearly correlated. 
Specifically, we first vectorize each background matrix and stack them together as a single matrix as { \bf B} = [vec({ \bf$b_1$}), vec({ \bf$b_2$}), {$\dots$}, vec({ \bf$b_n$})]. Theoretically, this matrix should be approximately low-rank. Besides, we employ the sparse matrix to model the foreground cells as { \bf E} = [ vec({ \bf$e_1$}), vec({ \bf$e_2$}), {$\dots$}, vec({ \bf$e_n$})]. To capture the structures in cells, we integrate  generalized fused lasso (GFL)~\cite{DBLP:journals/corr/XinTWG15} in the model. Therefore, we formulate the problem as:
% In order to estimate the low rank matrix {\bf B}, a matrix decomposition objective function, called generalized fused lasso (GFL), has been proposed as in
\begin{equation}
\begin{aligned}
& \underset{B,E}{\text{min}}
& & rank({\bf B}) + \lambda \lVert {\bf E} \rVert _{gfl},
& & \text{s.t.} 
& & {\large \bf A = B + E, }
\end{aligned}
\end{equation}
where $\lambda$ is a positive trade-off parameter and the definition of ${\lVert \cdot \rVert _{gfl}}$ is written as: 
\begin{equation}
\begin{aligned}
& & \lVert {\bf E} \rVert _{gfl} =  \sum_{k=1}^n \{ \lVert e_{k} \rVert _{1}   + \gamma \underset{(p,q) \varepsilon N} \sum \omega^{pq} _{k}  \lvert e^{p} _{k} -e^{q} _{k} \rvert  \},
\end{aligned}
\end{equation}
where $e_{k}$ is the foreground of $\emph{k-}th$ frame in an image sequence; the pixel $p$ and $q$ are the spatial neighborhood in the set $N$ (i.e. 4-connected neighborhood); $\gamma$ is a heuristic parameter, which is used to balance the sparsity and structural information of objects; the $\omega^{pq} _{k}$ is weights between pixel $\emph{p}$ and $\emph{q}$ and computed as 
$\omega^{pq} _{k} = e^{\frac{- \lVert I^{p} _{k} -I^{q} _{k} \rVert ^2 _2} {2 \sigma ^{2}}}$.
%When set $\gamma =0 $, it'll go back to the Robust Principle Analysis (RPCA) model~\cite{NIPS2009_3704}. 
% To capture the structural information of the foreground, ${ \omega^{pq}_k }$ is introduced by measuring the intensity similarity within a given neighbourhood, which is defined by
% \begin{equation}
% \begin{aligned}
% & & \omega^{pq} _{k} = e^{\frac{- \lVert I^{p} _{k} -I^{q} _{k} \rVert ^2 _2} {2 \sigma ^{2}}},
% \end{aligned}
% \end{equation}
% where $I^{p} _{k}$ and $I^{q} _{k}$ are the intensity of pixel $\emph{p}$ and $\emph{q}$ of the $\emph{k-}th$ image; $\sigma$ is a heuristic parameter. 
% Due to the special structure of cells in phase contrast images, we proposed a new phase retardation based weight and further explanation will be given in section
% 2.2.2. 
However, the rank function is hard to be optimized due to its non-convexity. We then use the nuclear norm, which is the sum of the singular values of ${\bf B}$, as an alternative relax solution: 
\begin{equation}\label{obj}
\begin{aligned}
& \underset{B,E}{\text{min}}
& & \lVert {\bf B} \rVert _{\ast} + \lambda \lVert {\bf E} \rVert _{gfl},
& \text{s.t.}
& & {\large \bf A = B + E},
\end{aligned}
\end{equation}
where the ${ \lVert \cdot \rVert _{\ast}}$ means the nuclear norm of a matrix, such as the sum of the matrix's singular values.

\begin{figure*}
\centering
\includegraphics[width=6.8in,height=0.6in]{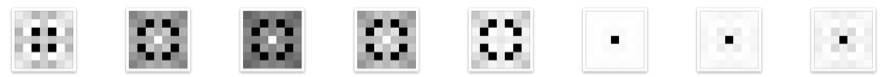}
\caption{ One inverse diffraction pattern filter bank with 8 phases. From left to right are   $\theta_1=0$, ${\theta_2=\pi/4}$,${\cdots}$, ${\theta_8=7\pi/4}$.}
\label{fig:short}
\end{figure*}

\begin{figure*}
\centering
\includegraphics[width=.255\textwidth]{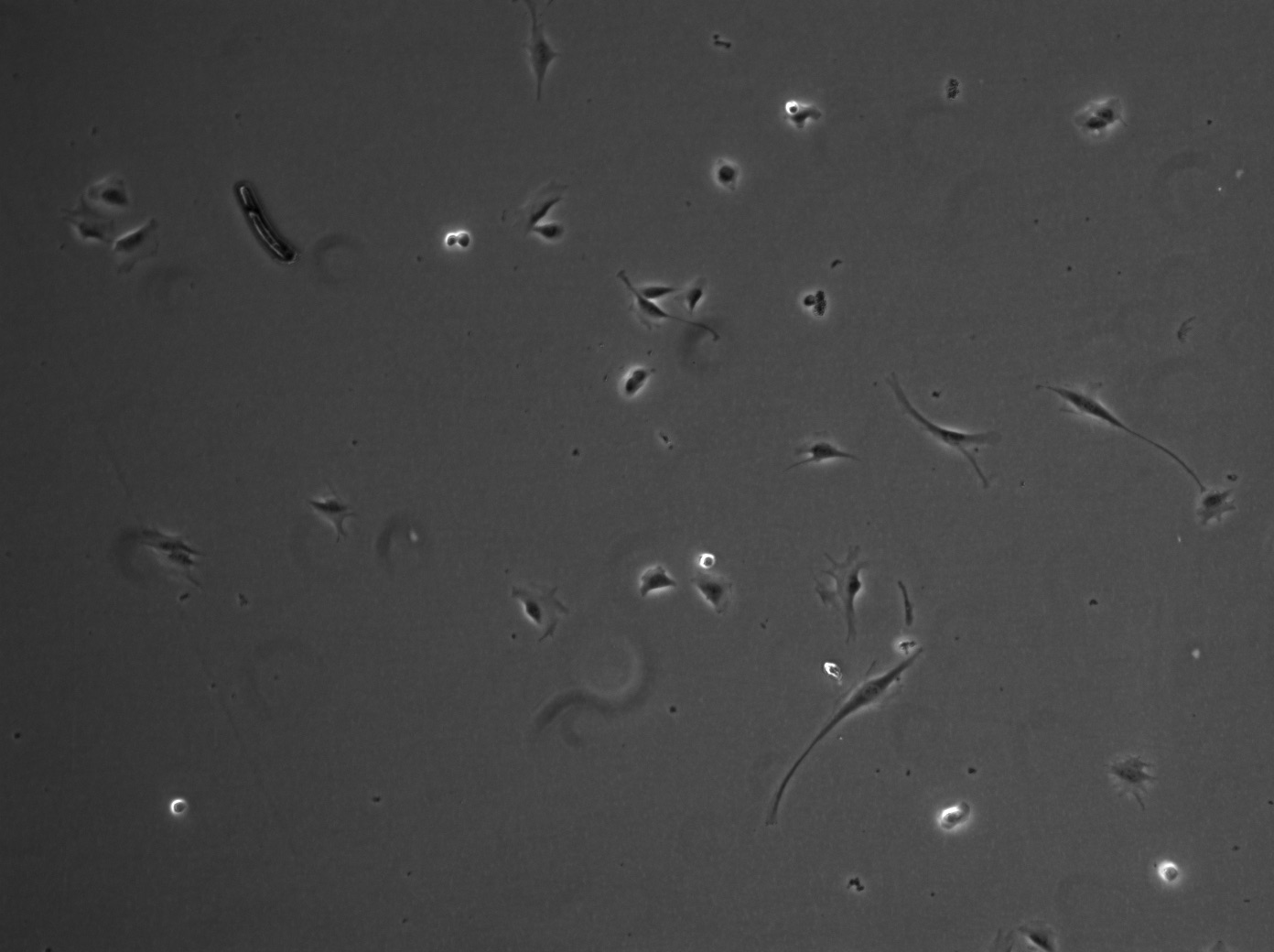}
\includegraphics[width=0.74\textwidth]{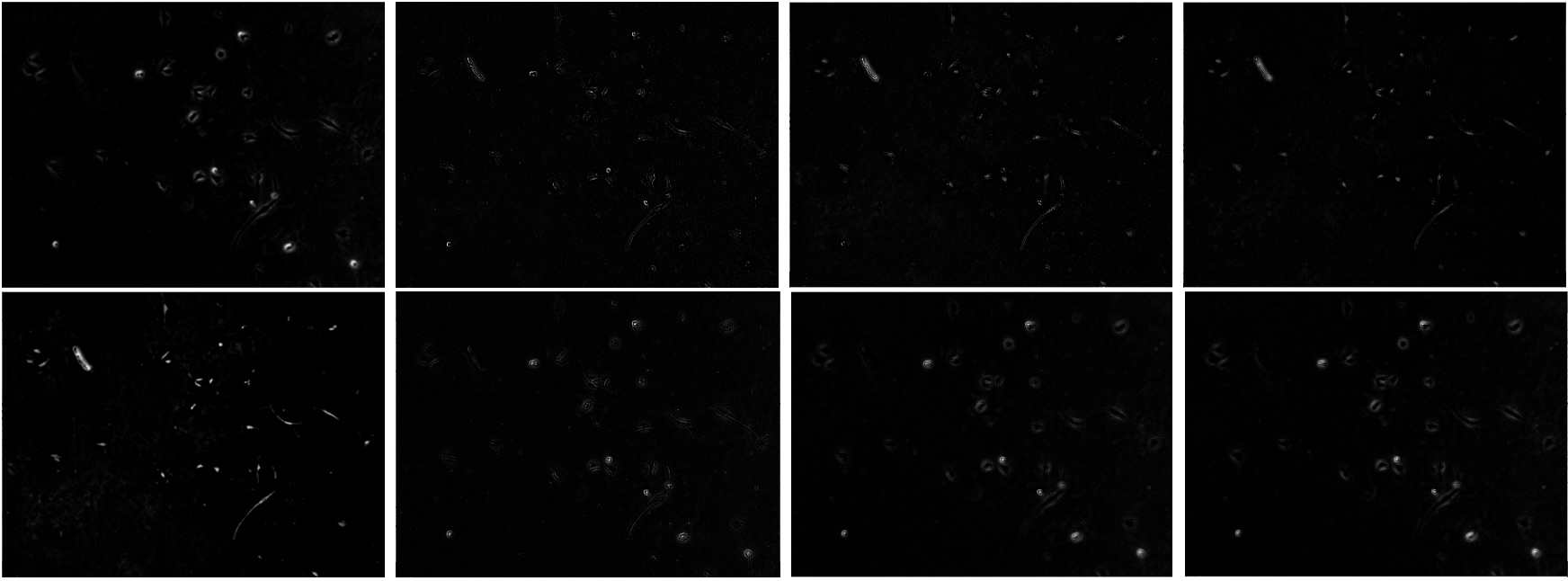}

\caption{An input phase contrast image(Leftmost); Eight filtered outputs by applying filters on the image after the background subtraction(upper row are ${\Phi_{1},\cdots,\Phi_{4}}$; bottom row are ${\Phi_{5},\cdots,\Phi_{8}}$.)}
\end{figure*}

\subsubsection{Optimization}
We now illustrate how to optimize Eq.~\ref{obj} based on the Augmented Lagrange Multipliers (ALM)~\cite{5540138,ALM}. We introduce the Lagrange multiplier ${\large \bf Y}$, the problem can be rewritten as
% In, an optimization solution based on the Augmented Lagrange Multipliers (ALM) is proposed to solve the convex problem in Eq.4 as
\begin{equation}
\begin{aligned}
& & L({\large \bf B,E},\lambda) =  \lVert  {\large \bf B} \rVert _{\ast} + \lambda \lVert {\large \bf E} \rVert _{gfl} + \langle {\large \bf Y, A-B-E} \rangle \\
& &  + \frac{\mu}{2} \lVert  {\large \bf A-B-E} \rVert ^2 _{F},
\end{aligned}
\end{equation}
where $\mu$ is a positive scalar.
%Here ${\lVert \cdot \rVert _{{\large \bf F}}}$ is the Frobenius norm. 
The optimal solution of ${\large \bf B}$ and ${\large \bf B}$ can be computed in an alternative way.
Firstly, we update the ${ \bf B_{t+1}}$ by fixing ${ \bf E} =  { \bf E} _{t}$,
\begin{equation}
\begin{aligned}
& {\large \bf B}_{t+1} =  \arg\min_{{\large \bf B}} L( {\large \bf B}, {\large \bf E}_{t}, {\large \bf Y}_{t},\mu_{t}) \\
& =\arg\min_{{\large \bf B}} \lVert  {\large \bf B} \rVert _{\ast} + \langle {{\large \bf Y}_{t}, {\large \bf A}-{\large \bf B}-{\large \bf E}_{t}} \rangle \\
& + \frac{\mu_{t}}{2} \lVert  {\large \bf A-B}-{\large \bf E}_{t} \rVert ^2 _{F},
\end{aligned}
\end{equation}
Next, we update { \bf E} by fixed the ${ \bf B}$ as: % _{t+1}$,
\begin{equation}
\begin{aligned}
& {\large \bf E} _{t+1} =  \arg\min_{\large \bf E} L({\large \bf B}_{t+1}, {\large \bf E}, {\large \bf Y}_{t},\mu_{t})\\
& = \arg\min_{\large \bf E}  \lambda \lVert {\large \bf E} \rVert _{gfl} + \langle {\large \bf Y}_t, {\large \bf A}-{\large \bf B}_{t+1}-{\large \bf E} \rangle \\
& +\frac{\mu_{t}}{2} \lVert  {\large \bf A}-{\large \bf B}_{t+1}-{\large \bf E} \rVert ^2 _{F},
\end{aligned}
\end{equation}
% When ${ \bf B}_{t+1}$ and ${\large \bf E}_{t+1}$ are converged and renamed as ${{\large \bf B}_{t+1}}$ and ${{\large \bf E}^{\ast}_{t+1}}$ respectively, update the optimal Lagrange multiplier ${\large \bf Y}_{t}$ as
In each iteration, the Lagrange multiplier is updated as:
\begin{equation}
\begin{aligned}
&& {\large \bf Y}_{t+1} = {\large \bf Y}_{t}+\mu_t ({\large \bf A}-{\large \bf B}_{t+1}-{\large \bf E}_{t+1}),
\end{aligned}
\end{equation}
Meanwhile, the parameter $\mu_{t}$ is updated accordingly as:
\begin{equation}
\begin{aligned}
\mu_{t+1} = 
\begin{cases}
{\rho \mu_t},\textup{if} \ \mu_t \lVert {\large \bf E}_{t+1}^{\ast}-{\large \bf E}_{t}^{\ast} \rVert _{F}  /  {\lVert {\large \bf A} \rVert} _{F} < \varepsilon  \\
\mu_{t}, \ \textup{otherwise}
\end{cases} 
\end{aligned}
\end{equation}
where $\rho$ is a constant that is larger than $1$;  $\varepsilon$ is a very small positive scalar. 
%More details of this algorithm can be found in ~\cite{5540138,ALM}.

%-------------------------------------------------------------------------
\subsection{Inverse Diffraction Pattern(IDP) filtering}
% We propose inverse diffraction pattern (IDP) filtering to restore artifacts-free phase contrast microscopy images efficiently. 
% %Based on this, we introduce a new way to obtain structural prior information.  
% \subsubsection{The diffraction pattern of phase contrast microscopy images} 
In~\cite{Zhaozheng}, the phase contrast microscopy image is approximated by a linear combination of $M$ diffraction patterns as:
% , which is due to the different refractive indices between cells and medium,
\begin{equation}
\begin{aligned}
% \!g\!=\!c\!+\!d\!\sum_{m=1}^M\!\Phi_m\!\ast\![sin\theta_m \delta(r)\!+\!(\varsigma_p cos\theta_m\!-\!sin\theta_m) airy(r)]\\
% & =c + d\sum_{m=1}^M \Phi_m \ast PSF(\theta_m),
% %\qquad \qquad \qquad \qquad \qquad  (5)
& \!g\!\propto\!\sum_{m=1}^M\!\Phi_m\!\ast\![sin\theta_m \delta(r)\!+\!(\varsigma_p cos\theta_m\!-\!sin\theta_m) airy(r)],
\end{aligned}
\end{equation}
%  are equally divided as  $2\pi$ like
where ${\theta_m}$ denotes the $\emph{m-}th$ phase retardation, which is defined as  $\theta_m =  2\pi \frac{m-1}{M}$; $\delta (r)$ is the Dirac delta function; $\varsigma_p$ is the amplitude attenuation factor caused by the phase ring, and we treat it as a constant at here;  $\Phi_m$ is the coefficient of $\emph{m-}th$ basis; $airy(r)$ is an obscured airy pattern with the radius $r$ that is defined in~\cite{Zhaozheng}.
% \begin{equation}
% \begin{aligned}
% airy(r) = R\frac{J_1(2 \pi Rr)}{r} - (R-W)\frac{J_1(2 \pi (R-W)r)}{r},
% \end{aligned}
% \end{equation}
% where \emph{R} and \emph{W} are the outer radius and width of the phase ring respectively. $J_1(\cdot)$ denotes the first order of Bessel function. 
For simplification, we define that 
\begin{equation}
\begin{aligned}
&  PSF(\theta_m) = sin\theta_m \delta(r)\!+\!(\varsigma_p cos\theta_m\!-\!sin\theta_m) airy(r),
\end{aligned}
\end{equation}
where $PSF(\theta_m)$ is the point spread function of the phase retardation $\theta_m$.

Mathematically, this is convolution process that images convolve with a complicated kernel. In order to restore the ideal image, we need deconvolution process that reverses the effects of the convolution on the observed data. 
More specifically, we need to seek the solution of a convolution equation of the form: 
\begin{equation}
\begin{aligned}
   h \ast f  = y, 
\end{aligned}
\end{equation}
where $ \emph{h} $ is the ideal image and contaminated by convolving with the kernel $ \emph{f}$. The convolution result, which is the observed image, is $ \emph{y}$. 
To serve such purpose, we apply deconvolution in the Frequency domain, which is written as:
\begin{equation}
\begin{aligned}
&  HF = Y,
\end{aligned}
\end{equation}
where $\emph{Y}$, $\emph{H}$ and $\emph{F}$ denote the Fourier transformation of $\emph{y}$, $\emph{h}$ and $\emph{f}$ respectively. 
By computing the inverse filter of ${\emph{F}}$ as
\begin{equation}
\begin{aligned}
 {F^{-1}}  = \frac{H}{Y},
\end{aligned}
\end{equation}
we can restore the ${ H}$ as  ${F^{-1} F H = H }$.
Consequently, the inverse filter can be computed in the time domain as
\begin{equation}
\begin{aligned}
 {f^{-1}}  = \mathfrak{F}^{-1}(F^{-1}),
\end{aligned}
\end{equation}
where ${f^{-1}}$ is the inverse filter of $f$; $\mathfrak{F}^{-1}$ is the inverse Fourier transform. With the definition of $PSF^{-1}(\theta_m)$ as the inverse of the $PSF(\theta_m)$, which is called as $\emph{inverse diffraction  pattern (IDP)}$ and shown in Figure 2, we can recover ideal images as:
\begin{equation}
\begin{aligned}
 \Phi_m = \bar{g} \ast PSF^{-1}(\theta_m),
\end{aligned}
\end{equation}
where $\bar{g}$ is the phase contrast microscopy image after background subtraction by the proposed method.

Usually, simple deconvolution is not robust due to the influence of noise, especially artifacts. In the presence of non-negligible noise, noise amplification will cause severe distortion~\cite{noise}. The reason makes it success in our task is that noise is suppressed successfully by the proposed background subtraction method. Since the imaging condition is maintained to be unchanged for a certain sequence, the noise is also similar in each frame. Thus, the noise among frames are also approximately linearly correlated, which is similar to the medium. Thus, it can be treated as a part of the background and removed by the proposed method efficiently. 

%-------------------------------------------------------------------------

%-------------------------------------------------------------------------

\section{Experimental Results}
% To validate the effectiveness of the proposed method, we apply the proposed background subtraction method and inverse diffraction pattern filtering method on cell segmentation.

\begin{table}
\begin{center}
%\begin{tabular}{|l|c|c|c|c|}
\begin{tabular}{|l|c|c|c|}
\hline
 & Dish1 & Dish2 & Dish3 \\
\hline
Otsu threshold~\cite{Otsu}   & 0.677 &0.665& 0.628\\ 
% \hline
% Preconditioning~\cite{understanding} & 0.974 & 0.974& 0.956 \\
% \hline
% Preconditioning~\cite{understanding} & 0.974 & 0.974& 0.956\\
\hline
RPCA~\cite{DBLP:journals/corr/XinTWG15} + IDP & 0.964 & 0.943 & 0.930\\
\hline
Cell-sensitive~\cite{cellsensitive} &  0.993 & 0.994 & 0.975\\
\hline
{Ours} & 0.995 & 0.995 & 0.978\\
\hline
\end{tabular}
\end{center}
\caption{The comparison of cell segmentation ACC on three dishes.}
\end{table}

\subsection{Evaluation Metrics and Data}

\begin{table}
\begin{center}
\begin{tabular}{|l|c|c|c|c|}
\hline
 & Dish1 & Dish2 & Dish3 & Time\\
\hline
Preconditioning~\cite{understanding} & 0.974 & 0.974& 0.956 & 262 sec \\
\hline
{Ours} & 0.995 & 0.995 & 0.981 & $<$ 1 sec\\
\hline
\end{tabular}
\end{center}
\caption{Performance comparison between ours and the Preconditioning restoration method.}
\end{table}

\begin{figure*}
\centering
\includegraphics[width=0.8\textwidth]{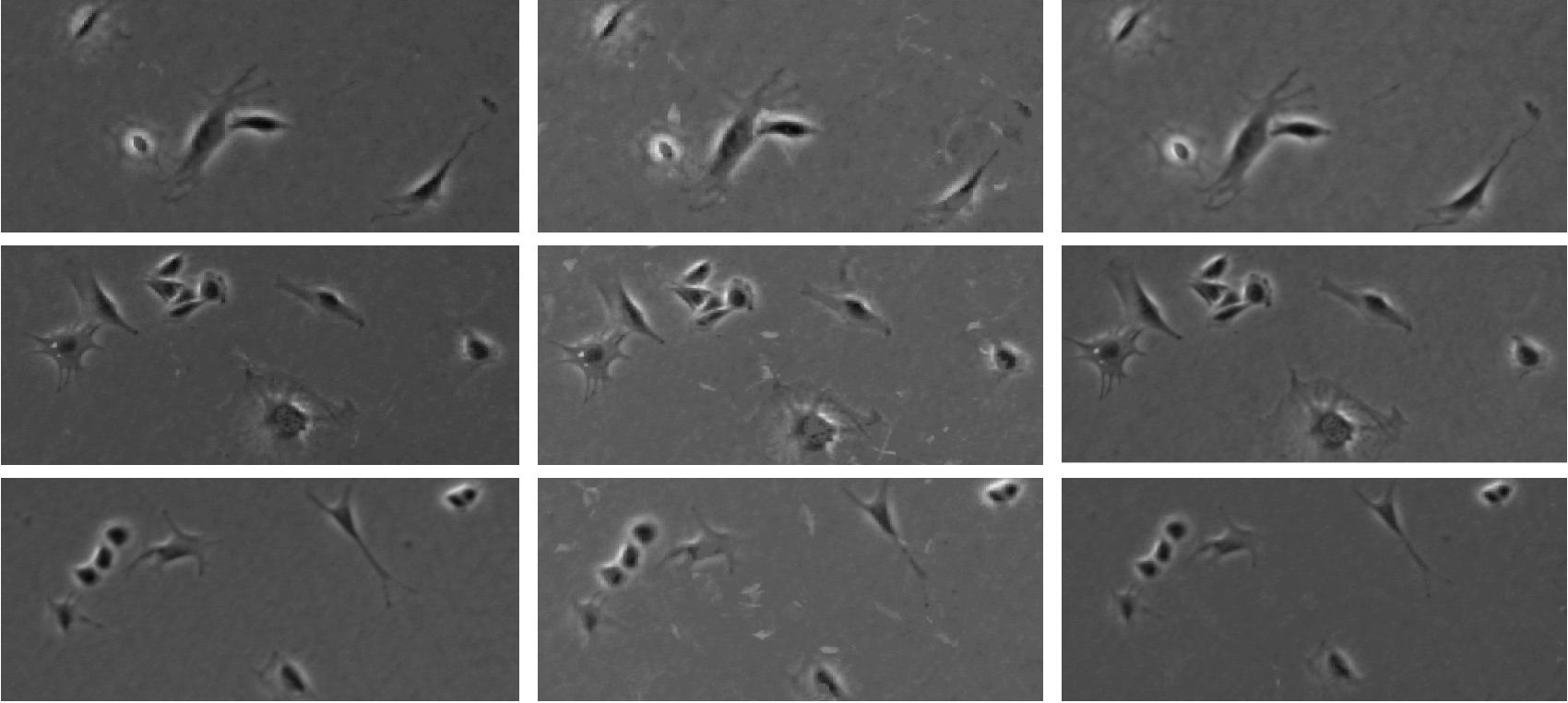}

\caption{Left: input images; Middle: the background subtraction results by RPCA; Right: the background subtraction results by ours}
\end{figure*}

The evaluation metric and data in~\cite{cellsensitive} are adopted in the experiments to facilitate comparisons with related works. The definition of the evaluation metric, accuracy, is $ACC = (\lvert TP\rvert + \lvert N \rvert-\lvert FP \rvert )/( \lvert P \rvert + \vert N \rvert)$, where $P$ denotes the positives, $N$ denotes the negatives, $TP$ denotes the true positives and $FP$ denotes the false positives. The data set consists of three different cell dishes under different exposure durations and the ground truth data is labeled by two annotators. 
% ([50 100 200 300 400 500]ms)
\subsection{Comparison and Evaluation}
We conduct the performance comparison with different methods, shown in Table 1. %The proposed method achieves a significantly better result than the Otsu threshold~\cite{Otsu}. 
Both the preconditioning method~\cite{understanding} and the proposed method achieved satisfactory results, where ours is slightly better than the preconditioning method. Our speed is much faster than the preconditioning method on these datasets, shown in Table 2. 
% To quantitatively compare with GFL model, we apply IDP filtering after its background subtraction. The proposed method didn't achieve significantly better results on cell segmentation than that of the GFL model, but our method has better performance on preserving the structure of cells than GFL, shown in Figure 3.
Although the performance of the proposed method is comparable to the cell-sensitive imaging~\cite{cellsensitive}, the proposed method holds some advantages. In cell-sensitive imaging method, the exposure time of images is a required parameter. Since the imaging information is not always available, this limits its application. Our method is designed to be independent of such parameters. This makes the proposed method more generally applicable. Also, our results demonstrate that the proposed method has better performance on removing  artifacts in background images when compared to the cell-sensitive imaging method, shown in Figure 4. For qualitative evaluation, Figure 5 shows more results on background subtraction and restored artifact-free phase contrast images by the proposed method. 
%But our method has a significantly advantage in the computation because the preconditioning is solved by a regularized quadratic cost function, which the computation is much heavier than ours%.

\begin{figure*}
\centering
\includegraphics[width=0.8\textwidth]{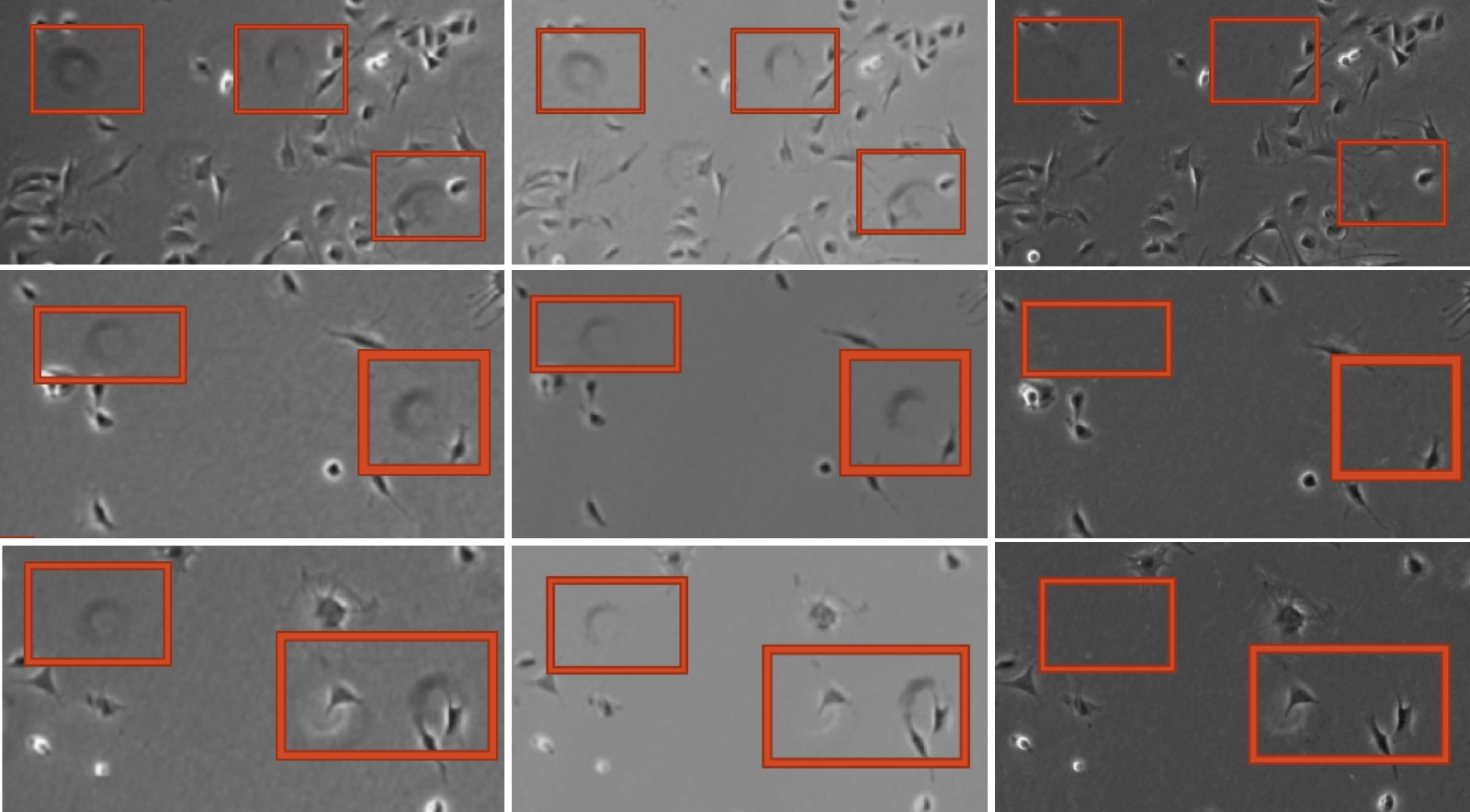}
\caption{Left: input images with artifacts(red box); Middle: the results by cell-sensitive imaging method; Right: the results by the proposed background subtraction method}
\end{figure*}

%  \begin{figure}
% \centering
% \includegraphics[width=2.5in,height=2.4in]{rings_2.jpg}
% \caption{Left column: raw data with artifacts(orange box); Right column: artifacts removed clearly after background subtraction.}
% \end{figure}

%--------------------------------------------
\begin{figure*}
% \centering
% \includegraphics[width=.3\textwidth]{Picture2.jpg}
% \includegraphics[width=.3\textwidth]{Picture3.jpg}
% \includegraphics[width=.3\textwidth]{Picture21.jpg}

\centering
\includegraphics[width=.3\textwidth]{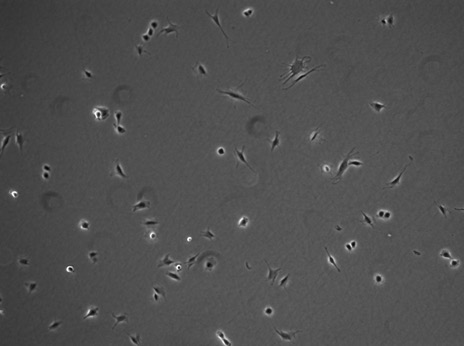}
\includegraphics[width=.3\textwidth]{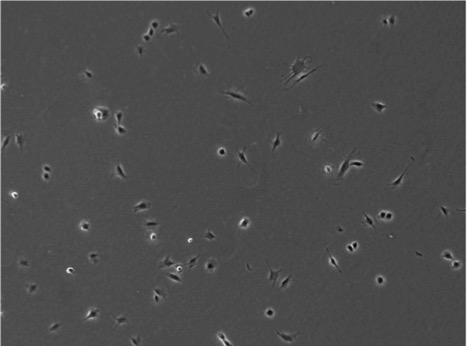}
\includegraphics[width=.3\textwidth]{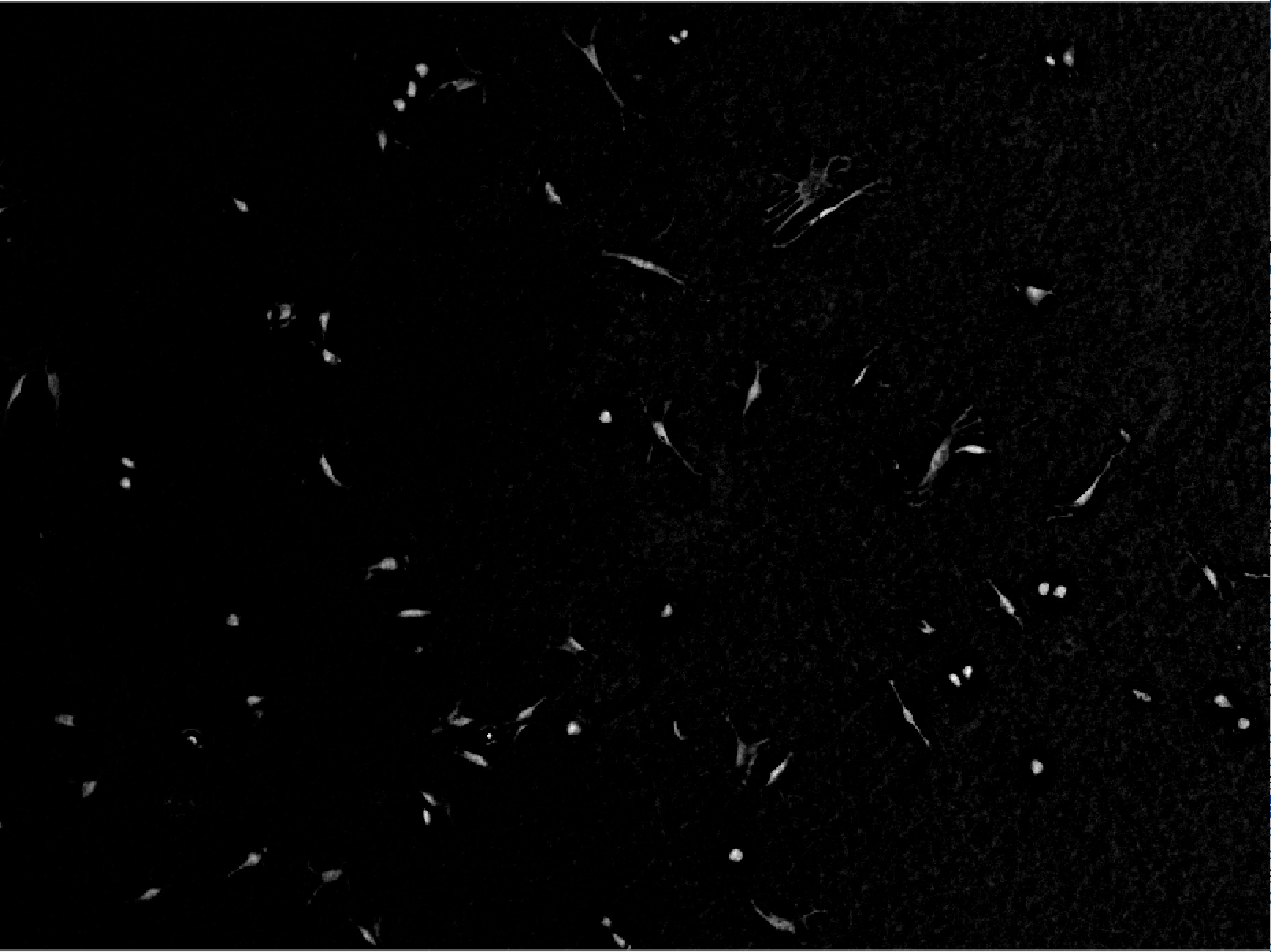}

\centering
\includegraphics[width=.3\textwidth]{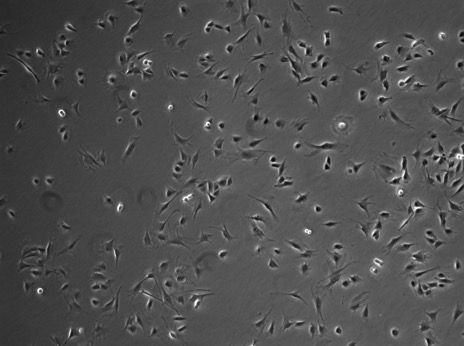}
\includegraphics[width=.3\textwidth]{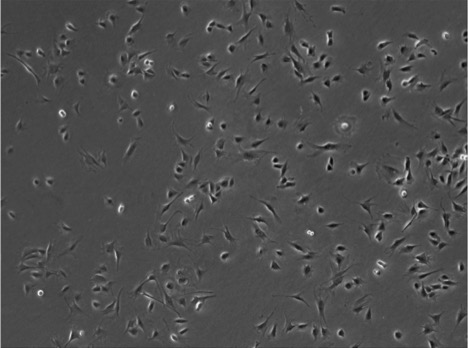}
\includegraphics[width=.3\textwidth]{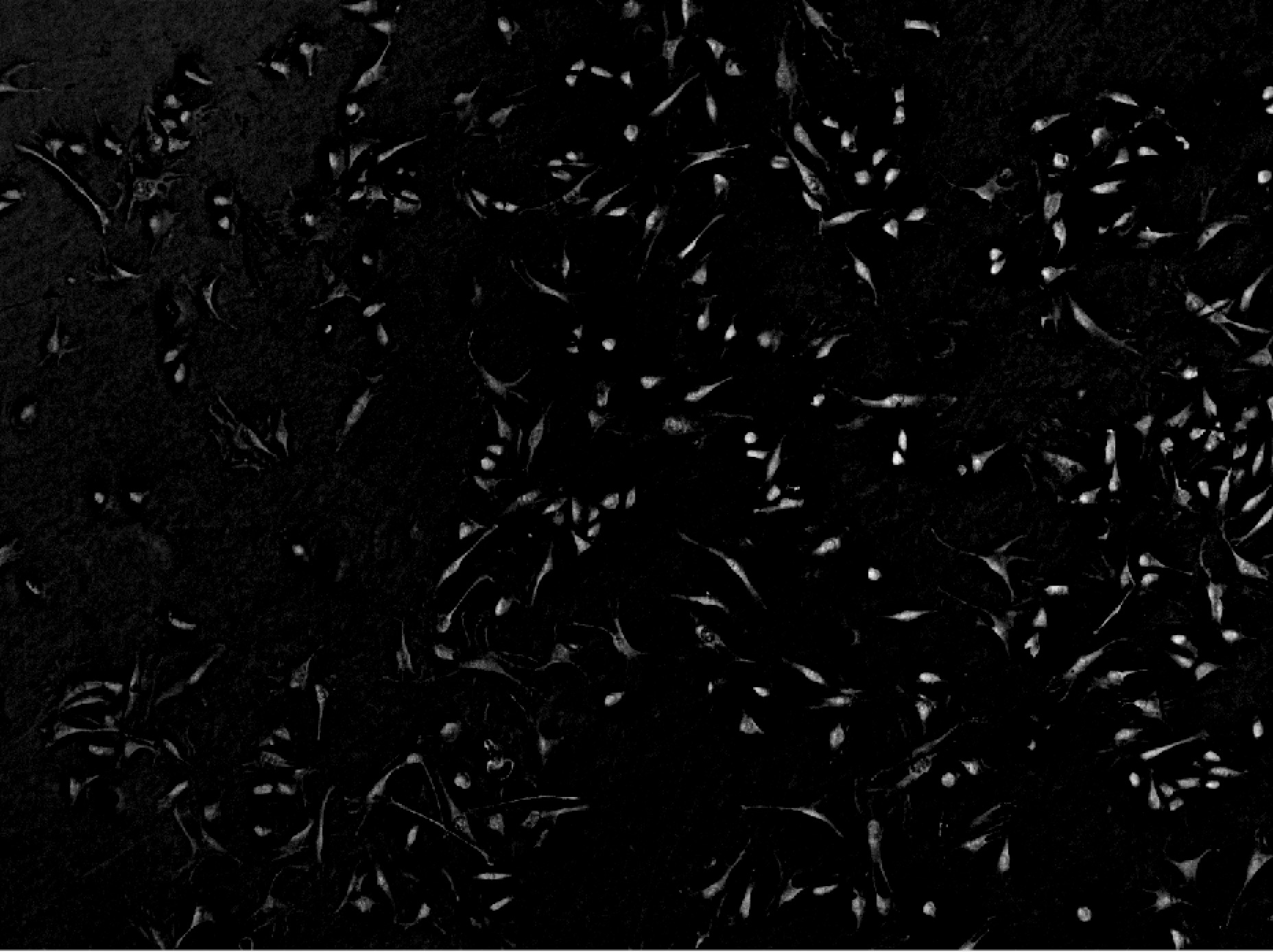}

\caption{Column 1: input phase contrast images; Column 2: images after background subtraction; Column 3: restored phase contrast images by deconvolution;}
\end{figure*}

%--------------------------------------------

\section{Conclusion}
In this paper, we propose a new approach for cell segmentation with two stages. We first remove background through a low-rank and sparse matrix decomposition. 
%To better keep the structure of cells, we introduce a structure constraint that is specially designed for phase contrast microscopy images. 
Then we obtain the accurate cells by introducing the inverse diffraction pattern filtering. It is inspired by optics model based restoration methods but much more efficient than them. Our experiments validate the effectiveness of the proposed method on cell segmentation. %Future work would be to combine with more prior information into the framework for more accurate segmentation. 

%\section{Acknowledgements}
% This research was supported/partially supported by [Name of Foundation, Grant maker, Donor]. We thank our colleagues from [Name of the supporting institution] who provided insight and expertise that greatly assisted the research, although they may not agree with all of the interpretations/conclusions of this paper.

% We thank [Name Surname, title] for assistance with [particular technique, methodology], and [Name Surname, position, institution name] for comments that greatly improved the manuscript.

% We would also like to show our gratitude to the (Name Surname, title, institution) for sharing their pearls of wisdom with us during the course of this research, and we thank 3 “anonymous” reviewers for their so-called insights. We are also immensely grateful to (List names and positions) for their comments on an earlier version of the manuscript, although any errors are our own and should not tarnish the reputations of these esteemed persons.

% We are immensely grateful to Dr.Seungil Huh from Google Inc. for sharing the idea of diffraction pattern filtering with us, where we got inspiration of inverse diffraction pattern filtering.
% We thank Dr.Zhaozheng Yin from Missouri University of Science and Technology, and his PhD student Wenchao Jiang for sharing their data with us. 
% We would also like to show our gratitude to Dr.Elmer Ker from Stanford University who provided data, insight, and expertise from biological perspective that greatly assisted our research.

{\small
\bibliographystyle{ieee}
\bibliography{egbib}

\begin{thebibliography}{10}\itemsep=-1pt

\bibitem{conf/miccai/FunkeHZ15}
J.~Funke, F.~A. Hamprecht, and C.~Zhang.
\newblock Learning to segment: Training hierarchical segmentation under a
  topological loss.
\newblock In N.~Navab, J.~Hornegger, W.~M.~W. III, and A.~F. Frangi, editors,
  {\em MICCAI (3)}, volume 9351 of {\em Lecture Notes in Computer Science},
  pages 268--275. Springer, 2015.

\bibitem{Li2007}
K.~Li, M.~Chen, and T.~Kanade.
\newblock {\em Medical Image Computing and Computer-Assisted Intervention --
  MICCAI 2007: 10th International Conference, Brisbane, Australia, October 29 -
  November 2, 2007, Proceedings, Part II}, chapter Cell Population Tracking and
  Lineage Construction with Spatiotemporal Context, pages 295--302.
\newblock Springer Berlin Heidelberg, Berlin, Heidelberg, 2007.

\bibitem{Snakes}
D.~T. Michael~Kass, Andrew~Witkin.
\newblock Snakes: active contour models.
\newblock {\em International Journal of Computer Vision}, 1:pp 321--331, 1988.

\bibitem{noise}
A.~M.R.Banhamand.
\newblock Digital image restoration.
\newblock {\em Signal Processing Magazine, IEEE}, 1:14(2):24–41., 1997.

\bibitem{Otsu}
N.~Otsu.
\newblock A threshold selection method from gray-level histograms.
\newblock {\em IEEE Trans. Syst., Man, Cybernet}, page 62–66, 1979.

\bibitem{5540037}
J.~Pan, T.~Kanade, and M.~Chen.
\newblock Heterogeneous conditional random field: Realizing joint detection and
  segmentation of cell regions in microscopic images.
\newblock In {\em Computer Vision and Pattern Recognition (CVPR), 2010 IEEE
  Conference on}, pages 2940--2947, June 2010.

\bibitem{5540138}
Y.~Peng, A.~Ganesh, J.~Wright, W.~Xu, and Y.~Ma.
\newblock Rasl: Robust alignment by sparse and low-rank decomposition for
  linearly correlated images.
\newblock pages 763--770, June 2010.

\bibitem{Zhaozheng}
H.~Su, Z.~Yin, T.~Kanade, and S.~Huh.
\newblock Phase contrast image restoration via dictionary representation of
  diffraction patterns.
\newblock pages 615--622, 2012.

\bibitem{NIPS2009_3704}
J.~Wright, A.~Ganesh, S.~Rao, Y.~Peng, and Y.~Ma.
\newblock Robust principal component analysis: Exact recovery of corrupted
  low-rank matrices via convex optimization.
\newblock pages 2080--2088, 2009.

\bibitem{DBLP:journals/corr/XinTWG15}
B.~Xin, Y.~Tian, Y.~Wang, and W.~Gao.
\newblock Background subtraction via generalized fused lasso foreground
  modeling.
\newblock {\em CoRR}, abs/1504.03707, 2015.

\bibitem{understanding}
Z.~Yin, T.~Kanade, and M.~Chen.
\newblock Understanding the phase contrast optics to restore artifact-free
  microscopy images for segmentation.
\newblock {\em Medical Image Analysis}, 16(5):1047--1062, 2012.

\bibitem{ALM}
M.~C. Z.~Lin and Y.~Ma.
\newblock The augmented lagrange multiplier method for exact recovery of
  corrupted low-rank matrices.
\newblock {\em arXiv preprint arXiv:1009.5055,}, 2010.

\bibitem{cellsensitive}
E.~K. M.~L. Zhaozheng~Yin, Hang~Su and H.~Li.
\newblock Cell sensitive phase contrast microscopy imaging by multiple
  exposures.
\newblock {\em Medical Image Analysis (MedIA)}, 2015.

\end{thebibliography}
}

\end{document}